\documentclass[runningheads]{llncs}

\usepackage{eccv}

\usepackage{eccvabbrv}
\usepackage{graphicx}
\usepackage{booktabs}
\usepackage{multirow}
\usepackage{graphicx}
\usepackage{amsmath}
\usepackage{amssymb}
\usepackage{booktabs}
\usepackage{xcolor}
\usepackage{bm}
\usepackage{pifont}
\usepackage{enumitem}
\usepackage{soul}
\usepackage{colortbl}
\usepackage{dblfloatfix}
\usepackage{siunitx}
\usepackage{balance}
\usepackage{rotating}
\usepackage{makecell}
\usepackage{caption}
\usepackage{multicol}
\usepackage{wrapfig}
\usepackage[accsupp]{axessibility}  

\usepackage{hyperref}

\usepackage{orcidlink}

\newcommand{\tit}[1]{\smallbreak\noindent\textbf{#1.}}

\begin{document}

\title{Alfie:Democratising RGBA Image Generation With No \$\$\$\vspace{-0.2cm}}

\titlerunning{Alfie: RGBA Image Generation With No \$\$\$}

\author{Fabio Quattrini\orcidlink{0009-0004-3244-6186} \and Vittorio Pippi\orcidlink{0009-0001-7365-6348} \and \\ Silvia Cascianelli\orcidlink{0000-0001-7885-6050} \and Rita Cucchiara\orcidlink{0000-0002-2239-283X}}

\authorrunning{F.~Quattrini et al.}

\institute{University of Modena and Reggio Emilia, Modena, Italy\\
\email{\{name.surname\}@unimore.it}\vspace{-0.5cm}}

\maketitle

\begin{abstract}
    Designs and artworks are ubiquitous across various creative fields, requiring graphic design skills and dedicated software to create compositions that include many graphical elements, such as logos, icons, symbols, and art scenes, which are integral to visual storytelling. Automating the generation of such visual elements improves graphic designers' productivity, democratizes and innovates the creative industry, and helps generate more realistic synthetic data for related tasks. These illustration elements are mostly RGBA images with irregular shapes and cutouts, facilitating blending and scene composition. However, most image generation models are incapable of generating such images and achieving this capability requires expensive computational resources, specific training recipes, or post-processing solutions. In this work, we propose a fully-automated approach for obtaining RGBA illustrations by modifying the inference-time behavior of a pre-trained Diffusion Transformer model, exploiting the prompt-guided controllability and visual quality offered by such models with no additional computational cost. We force the generation of entire subjects without sharp croppings, whose background is easily removed for seamless integration into design projects or artistic scenes. We show with a user study that, in most cases, users prefer our solution over generating and then matting an image, and we show that our generated illustrations yield good results when used as inputs for composite scene generation pipelines. We release the code at~\url{https://github.com/aimagelab/Alfie}. 

\end{abstract}

\section{Introduction}
\begin{figure}[t]
    \centering
    \includegraphics[width=\linewidth]{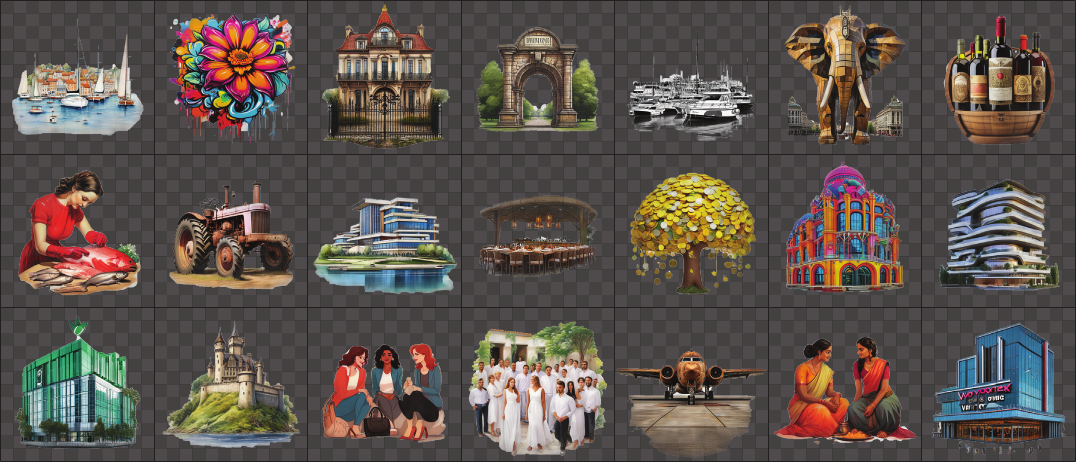}
    \vspace{-0.5cm}
    \caption{We propose a fully-automated pipeline to generate RGBA illustrations by adapting the inference-time behavior of a Diffusion Transformer model.}\vspace{-0.5cm}
    \label{fig:teaser}
\end{figure}

We are surrounded by a lot of visual content, including visually-rich documents in everyday life, such as pamphlets and banners, technical or scientific papers, business reports, storybooks, or novels with intricate and meaningful illustrations. Much effort has been dedicated to generating and integrating multiple design elements into visually-rich documents~\cite{zheng2019content,li2021towards,yamaguchi2021canvasvae,huang2023survey}.
Note that, when composing visually-rich documents (\eg~slides for a presentation, banners, cards, pamphlets, social media posts), it is not uncommon that designers, content creators, and even non-expert users would like to insert images that should integrate and blend nicely with the rest. Those users rely more and more often on AI-generated images for their designs. Generative models, especially prompt-driven diffusion models, have achieved impressive performance, and the increasing availability of user-friendly interfaces also eases non-experts' use of them. Moreover, these models allow for obtaining control over the generated content and producing images free from copyright. Nonetheless, when integrating an illustration inside a design, the designer often wants just the subject of interest to be superimposed on an existing background (possibly patterned or consisting of another image). 

Large-scale image generation models have been widely adopted and deeply impacted visual arts. Still, layered image generation has not been extensively tackled, mainly because of the lack of training data. While RGB image generation can leverage open datasets of billions of samples~\cite{schuhmann2022laion}, RGBA image datasets are of limited size~\cite{rhemann2009perceptually,xu2017deep,sun2021semantic}, the biggest being the recently proposed MAGICK~\cite{burgert2024magick} which contains only 150k elements. These datasets are commonly used to train matting models to estimate the alpha channel of a given image~\cite{yao2024matte,sun2021semantic,xu2017deep,hu2023diffusion,yao2024vitmatte}. Another solution to obtain RGBA images consists in directly predicting all four channels at once by fine-tuning a pre-trained generative model~\cite{zhang2024transparent}. However, these solutions are costly for the required training resources. 

To overcome these limitations, in this work, we propose \textit{Alfie}, a pipeline to obtain high-quality, prompt-driven illustrations to be seamlessly integrated into any design or document with limited to no editing effort by the user (see~\cref{fig:teaser}). 
We exploit a pre-trained Diffusion Transformer model~\cite{peebles2023scalable,chen2023pixartalpha}, \ie~the freely-available PixArt-$\Sigma$~\cite{chen2024pixartsigma}, and modify its diffusion process in two ways. First, we mask and combine two latents (one for the subject and one for the background) in order to force the generation of the subject in the center of the canvas and with no sharp crops so that these can be realistic in a design or document. Second, we use the informative cross- and self-attention maps~\cite{tang-etal-2023-daam,daam2024} computed during the generation process to extract the foreground regions and estimate the alpha channel values. 
We opt for a Diffusion Transformer for the greater flexibility of such models in generating images with different sizes and aspect ratios compared to standard U-Net-based diffusion models. This arguably makes them a more suitable starting point for the task at hand. Moreover, they are currently state-of-the-art both in generation quality and computational requirements~\cite{peebles2023scalable,chen2023pixartalpha,crowson2024scalable,chen2024pixartsigma}. PixArt-$\Sigma$~\cite{chen2024pixartsigma} (our baseline model) is composed of 0.6B parameters, compared to the 2.6B of Stable Diffusion XL~\cite{podell2023sdxl}. 
We quantitatively and qualitatively assess the capability of our approach to generate subjects that are fully contained within the image and its capability to adhere to the prompt. We analyze the cross- and self-attention maps computed during the generation process, exploring various alternatives for estimating the $\alpha$ (transparency) channel. Furthermore, we present a user study, which shows that users mostly prefer our method over generating and then matting (63\% of the time). Finally, we demonstrate the versatility of our generated illustrations by integrating them into a scene composition pipeline, achieving results of similar quality to those obtained by using illustrations from Adobe Stock. To facilitate further research on this task, we release the code of our approach and the evaluation setup\footnote{\url{https://github.com/aimagelab/Alfie}}.

\section{Related Work}
\label{sec:related}
\tit{Diffusion Models}
Since their introduction, diffusion models~\cite{sohl2015deep, song2019generative, song2020score, song2020improved, ho2020denoising, nichol2021improved, dhariwal2021diffusion} have been widely adopted for their impressive performance, 
even more in the text-to-image setting, where the generation is guided by a prompt in natural language (as for, \eg, DALL$\cdot$E 2~\cite{ramesh2022hierarchical}, SD-XL~\cite{podell2023sdxl}, and the PixArt family~\cite{chen2023pixartalpha, chen2024pixartsigma}). 
Such popularity has been further boosted by the introduction of latent diffusion models~\cite{rombach2022high}, which save computational resources in training and inference by working in the more compact latent space instead of the pixel space.
Moreover, recently proposed Diffusion Transformers~\cite{peebles2023scalable,chen2023pixartalpha,chen2024pixartsigma,nair2024diffscaler,crowson2024scalable} have further increased the scalability and obtainable visual quality of diffusion models by exploiting a fully-attentive Transformer model instead of the convolutional-attentive U-Net-like noise estimators adopted in previous works.
Lots of research efforts have also been dedicated to devising efficient fine-tuning strategies to obtain better controllability over the generation process of diffusion models based on additional guiding signals~\cite{gafni2022make, brooks2023instructpix2pix, avrahami2023spatext, zhang2023adding} while maintaining the visual quality of the output. Another line of research focuses on achieving zero-shot capabilities by proposing strategies to alter the inference-time behavior of pre-trained diffusion models~\cite{meng2021sdedit, lugmayr2022repaint, avrahami2023blended, mokady2023null, tumanyan2023plug}. Some of these approaches entail modifying the noisy latent vectors at each inference step. Some approaches perform mask-guided inpainting by combining the noised image with generated latent vectors~\cite{avrahami2023blended, lugmayr2022repaint}, or combine multiple independently generated latent vectors for region-based image generation~\cite{bar2023multidiffusion}. Other works exploit the gradient of a downstream task-related loss or score~\cite{clipguideddiffusion,lee2023syncdiffusion}, possibly computed on the foreseen image obtained from the latent vector. A line of work has focused on manipulating the cross- and self-attention activations for image editing~\cite{hertz2022prompt,couairon2022diffedit}, open-vocabulary segmentation~\cite{li2023open,nguyen2024dataset}, and zero-shot video generation~\cite{khachatryan2023text2video}. The role of attention layers in U-Net-based diffusion models has been extensively studied also from an explainability perspective by DAAM~\cite{tang-etal-2023-daam}. In this work, we use the cross-attention maps to identify the foreground pixels of the generated image and combine them with the self-attention maps to obtain the alpha channel.
In the context of generating designs and visually-rich documents, training, fine-tuning, and inference-time adaptation approaches for diffusion models have been proposed. These focus on generating aesthetically-pleasing document layouts~\cite{li2023relation,he2023diffusion,chen2023towards,chai2023layoutdm,inoue2023layoutdm}, layout-aware designs backgrounds~\cite{weng2024desigen}, user interfaces~\cite{hui2023unifying}, SVGs~\cite{jain2023vectorfusion,xing2024svgdreamer}, and collages~\cite{bar2023multidiffusion,zhang2023diffcollage,sarukkai2024collage}.
In this work, we propose to adapt the inference-time behavior of a Diffusion Transformer~\cite{chen2024pixartsigma} for a fully automated pipeline to obtain illustration-like images that are easy to add to any existing background. In particular, we explore the inference-time adaptation of such a model by combining multiple masked latent vectors to obtain the desired characteristics in the generated images.

\tit{RGBA Images} Several approaches exist to estimate the $\alpha$ channel of a given RGB image~\cite{zhu2015targeting}, which can be binary and indicate foreground and background only (as in segmentation strategies), or continuous between 0 and 1 to indicate the opacity of each pixel (as in from matting algorithms). Proposed approaches range from classical ones exploiting perceptual features of the image (such as color~\cite{ben2000segmentation} and edges~\cite{rother2004grabcut}) to more recent learning-based solutions~\cite{sun2021semantic,xu2017deep,hu2023diffusion,qin2022highly,kirillov2023segment,yao2024vitmatte}. Note that those approaches require some additional input to guide the prediction of the $\alpha$ channel. For example, image matting approaches~\cite{yao2024matte,sun2021semantic,xu2017deep,hu2023diffusion} need a map of foreground/background/unknown regions (a trimap), while even recent class-free segmentation approaches~\cite{lueddecke22image,kirillov2023segment,yao2024vitmatte} need anchor points or visual prompts indicating the subject of interest. 
In LayerDiffuse~\cite{zhang2024transparent}, the authors propose a fine-tuning strategy to enable transparent image generation using large-scale pre-trained diffusion models. In particular, they adjust the latent space of Stable Diffusion XL~\cite{podell2023sdxl} and finetune both the U-Net and the VAE, by using a loss that preserves the original latent distribution. This method, while effective, requires about 350 A100 hours for fine-tuning (as specified by the authors) and is not flexible to changes in the backbone model. In this work, we exploit the cross- and self-attention maps of our inference-time adapted Diffusion Transformer to obtain a guidance signal for generating RGBA images with satisfying results, especially when the output is used for further processing or scene composition.

\section{Preliminaries}
\setlength{\belowdisplayskip}{4pt} \setlength{\belowdisplayshortskip}{4pt}
\setlength{\abovedisplayskip}{4pt} \setlength{\abovedisplayshortskip}{4pt}

\tit{Diffusion Models}
By diffusion models~\cite{sohl2015deep, ho2020denoising, song2020score, song2020denoising} we refer to a class of generative probabilistic models trained to transform Gaussian noise $\mathbf{x}_T {\sim} \mathcal{N}(0, \mathbf{I})$ into samples belonging to a certain data distribution $\mathbf{x}_0 {\sim} q$ in $T$ steps. This is obtained by learning to approximate the data distribution $q$.
To this end, during the so-defined forward process, Gaussian noise is injected into the data to transform the data distribution into the marginal distribution, \ie
\begin{equation*}
    q(\mathbf{x}_t|\mathbf{x}_0) = \mathcal{N} (\mathbf{x}_t; \alpha_t\mathbf{x}_0, \sigma_t^2 \mathbf{I}),
\end{equation*}
where the parameters $(\alpha_t, \sigma_t)$ define a differentiable noise schedule ensuring that $q(\mathbf{x}_t){\approx} \mathcal{N}(0, \mathbf{I})$.
Then, the diffusion models framework entails a reverse process in which the model is trained to denoise $\mathbf{x}_T{\sim}q(\mathbf{x}_t|\mathbf{x}_0)$. To this end, $\mathbf{x}_0$ is predicted iteratively by estimating $\mathbf{x}_{t-1}$ starting from $\mathbf{x}_t$. The most common approach, proposed in~\cite{ho2020denoising}, entails using a parameterization based on the prediction of the noise $\mathbf{\epsilon}$ for sampling: both $\mathbf{x}_t$ and $\mathbf{x}_{t-1}$ are parametrized as a combination of $\mathbf{x}_0$ and $\mathbf{\epsilon}$, scheduled according to the noise schedule. As a result, the model is trained to estimate the noise $\mathbf{\epsilon}$ by optimizing 
\begin{equation*}
    \mathbb{E}_{q(\mathbf{x}_0)}[\|\mathbf{\epsilon}_{\theta}(\mathbf{x}_t, t) - \mathbf{\epsilon}\|_2^2].
\end{equation*}

\tit{Latent Diffusion Models} 
When working with high-resolution image data, training diffusion models in the pixel space is very computationally heavy. To face this issue, latent diffusion models~\cite{rombach2022high, podell2023sdxl} have been introduced. These models work with vectors in the latent space of an autoencoder (\eg~a VQ-VAE~\cite{van2017neural} or a VQ-GAN~\cite{esser2021taming}) and thus are more efficient than models working in the pixel-space in terms of GFlops~\cite{rombach2022high, podell2023sdxl, peebles2023scalable, chen2023pixartalpha}. Here, we rely on the pre-trained, latent Diffusion Transformer model PixArt-$\Sigma$~\cite{chen2024pixartsigma}.

\tit{Conditional Generation} 
The generation process of diffusion models can be conditioned on a guidance signal. The most popular diffusion models to date are text-to-image models that can be conditioned on a natural language prompt, embedded by the text encoder of a pre-trained multimodal model (\eg~CLIP~\cite{radford2021learning} or T5~\cite{raffel2020exploring}) in a vector $\mathbf{e}$. The conditioning is achieved by performing cross-attention with $\mathbf{e}$ inside the noise estimation network.  
Note that, in this work, we use as backbone a diffusion model trained with the Classifier-Free Guidance~\cite{ho2022classifier} conditioning strategy, in which the noise prediction $\hat{\mathbf{\epsilon}}_{\theta}$ is obtained by combining the conditional prediction $\mathbf{\epsilon}_{\theta}(\mathbf{x}_t, \mathbf{e}, t)$ and the unconditional prediction $\mathbf{\epsilon}_{\theta}(\mathbf{x}_t, \emptyset, t)$, where $\emptyset$ is the embedding vector of the null prompt, with weight $s$ as 
\begin{equation*}
    \hat{\mathbf{\epsilon}}_{\theta}(\mathbf{x}_t, \mathbf{e}, t) = \mathbf{\epsilon}_{\theta}(\mathbf{x}_t, \emptyset, t) + s(\mathbf{\epsilon}_{\theta}(\mathbf{x}_t, \mathbf{e}, t) - \mathbf{\epsilon}_{\theta}(\mathbf{x}_t, \mathbf{\emptyset}, t)).
\end{equation*}

\tit{Diffusion Transformers} 
Diffusion models originally featured a convolutional-attentive U-Net-like model as a noise estimator. The seminal work~\cite{peebles2023scalable} proposes to replace the U-Net with a multi-block Diffusion Transformer model (DiT) to achieve better performance and scalability. The noise estimation Transformer takes as input a sequence of tokens obtained from squared patches of the latent vectors added to a positional embedding. At each block of the Transformer, the generation is conditioned on the timestep and the class. After the last block, the tokens are decoded and rearranged into the final image spatial dimension.
Subsequent works~\cite{chen2023pixartalpha,chen2024pixartsigma} adapt the class-guided Diffusion Transformer to work with text guidance. In this work, we exploit the recently-proposed PixArt-$\Sigma$ Diffusion Transformer~\cite{chen2024pixartsigma}, which is directly conditioned on the textual prompt.\vspace{-2em}
\begin{figure}[t]
    \centering
    \includegraphics[width=\linewidth]{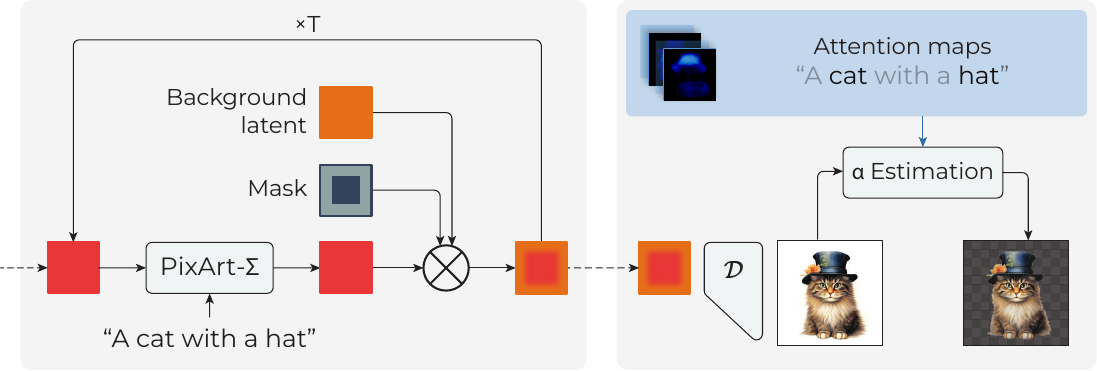}
    \vspace{-0.5cm}
    \caption{Schematic representation of our fully-automated prompt-guided pipeline to obtain RGBA illustrations. The core element is a diffusion model, for which we devise an inference-time adaptation strategy aimed at making the generated images illustration-like. Then, we process the generation attention maps to estimate the $\alpha$ channel.}\vspace{-0.5cm}
    \label{fig:pipeline}
\end{figure}

\section{Inference-Time Illustration Generation}\label{sec:method}
We propose a fully automated pipeline to obtain a high-quality RGBA illustration from a textual prompt. We argue that such illustration should have the following characteristics: 
\begin{enumerate}[noitemsep,topsep=0pt,leftmargin=*] 
    \item Contain the subjects specified in the prompt; 
    \item Be fully contained in the canvas without crops; 
    \item Have a precise $\alpha$ channel.
\end{enumerate}
In our proposed inference-time method, we tackle all three aspects. In particular, we exploit the generative power of a pre-trained DiT (1), altering its inference-time behavior so that it always generates entire subjects (2) and extracting and cleaning an $\alpha$ channel estimate (3). Our approach is summarized in~\cref{fig:pipeline}. 
We denote the RGB image as $\hat{x} \in \mathbb{R}^{h{\times}w{\times}3}$, with values in range $[-1, 1]$. The transparency channel is denoted as $\alpha \in \mathbb{R}^{h{\times}w{\times}1}$, with values in range $[0, 1]$. The complete RGBA image is obtained by concatenating the RGB image with its $\alpha$ channel and has shape $h{\times}w{\times}4$.

In particular, we propose to obtain the $\alpha$ channel by exploiting the relevant foreground subject's cross-attention maps and the pixels' self-attention map relative to the foreground subject. The rationale is that the cross-attention map will have higher values for the pixels of the subject than those of the background, providing a coarse localization of the subject. On the other hand, the self-attention values will help render the texture and appearance of the foreground subject. In fact, when a foreground pixel is relative to an object of solid material, it will attend mainly to neighboring pixels of the same object, resulting in a cohesive appearance. In this case, all the self-attention weights for the pixels of that object will be high.
Conversely, the appearance of foreground pixels of objects in a translucent, see-through material (\eg, water, fire, glass) will also depend on the background. Therefore, that pixel will attend to both the pixels of the translucent object and the background. As a result, the self-attention weights for the pixels of that object will be lower than in the case of solid objects. This behavior is in line with what the $\alpha$ channel represents. For this reason, we propose to use the cross-attention and self-attention maps to define the $\alpha$ channel of our generated RGBA illustrations.

\tit{Centering the Subject} 
Diffusion models are trained on billions of images~\cite{schuhmann2022laion} and learn to generate all kinds of styles, from natural to artworks. While some works analyze the disentanglement capability of such models~\cite{wu2023uncovering}, recent works argue that the low quality of the captions in most large-scale datasets negatively affects the generation capabilities and train on LLaVA~\cite{liu2023llava}-refined captions~\cite{chen2023pixartalpha, chen2024pixartsigma}. Still, complete disentangling has yet to be achieved and the models have difficulties with generating meta-descriptions or combinations of subjects and styles not seen in the training data. Note that by meta-descriptions, we define indications on the characteristics of the image, such as the background color or the fact that the full subject should be contained.

When generating illustrations, it is important that the subjects are wholly contained within the image canvas and are not cut or cropped. To achieve this, we chose to employ a method inspired by~\cite{avrahami2023blended, bar2023multidiffusion}, where we jointly generate the illustration (described by the textual prompt) and a uniform background, and we blend them together at each denoising step. Formally, to generate an image $\mathbf{x_0}$ with the text prompt $\mathbf{e}$, we first define a squared mask $\mathbf{m}$, covering the inner area of the canvas. Then, we start from the Gaussian noise $\mathbf{x}_t$ at the timestep $t \in [0,T]$ and duplicate it into a foreground latent $\mathbf{x}_{t, fg}$ and a background latent $\mathbf{x}_{t, bg}$, assigning a background textual prompt $\mathbf{e}_{bg}$ to the respective latent.
After the denoising step, we merge the obtained predictions into
\begin{equation*}
    \mathbf{x}_{t-1} = \mathbf{x}_{t-1, fg} \cdot \mathbf{m} + \mathbf{x}_{t-1, bg} \cdot (1 - \mathbf{m}).
\end{equation*}
This process constrains the illustration subjects to be generated inside the mask while ensuring that they are not cropped. After the denoising chain, the image is obtained from the final latent $\mathbf{x}_0$ by using the decoder $\mathbf{\hat{x}} = \mathcal{D}(\mathbf{x}_0)$.

\tit{Isolating the Subject}
During each forward process generation step $t$, each layer $l$ performs cross-attention between the latent $\mathbf{z}_{t,l}$ and the encoded prompt $\mathbf{e}$ to guide the subject creation and self-attention to correlate image features. Following DAAM~\cite{tang-etal-2023-daam} and DAAM-I2I~\cite{daam2024}, we extract the cross- and self-attention maps. Formally, for a given $t$ and $l$, the cross-attention map $\mathcal{A}_C^{t,l} \in [0,1]^{h{\times}w{\times}N}$ will represent the relation between each pixel and each of the $N$ prompt tokens, while the self-attention map $\mathcal{A}_S^{t,l} \in [0,1]^{hw{\times}h{\times}w}$ will represent the relation between each pixel to all the other latent pixels: 
\begin{align*}
    \mathcal{A}_C^{t,l} &:= \text{softmax} \left( Q_{\mathbf{z}_{t,l}} K_{\mathbf{e}}^T / \sqrt{d} \right), \\
    \mathcal{A}_S^{t,l} &:= \text{softmax} \left( Q_{\mathbf{z}_{t,l}} K_{\mathbf{z}_{t,l}}^T / \sqrt{d} \right),
\end{align*}
\noindent where $d$ is the feature size of the layer $l$. 

Note that by performing subject-centering and classifier-free guidance, we have a batch size of four, respectively representing the latent for the null ($\theta$) and text prompts ($\mathbf{e}$) for the background and the subject. We discard all the attention maps except those corresponding to the prompt-guided subject latent. We reserve further investigation on integrating the background and null-prompt guided attention maps for future work. Since diffusion models define the coarse image layout during the first steps of the denoising chain and later define the details, we keep the maps of the last ten out of thirty timesteps to obtain more precise localization. Finally, we average across timesteps, layers, and attention heads to obtain the global maps $\mathcal{A}_C \in [0,1]^{h{\times}w{\times}N}$ and $\mathcal{A}_S \in [0,1]^{hw{\times}h{\times}w}$.  

\begin{figure}[t]
    \centering
    \includegraphics[width=.75\linewidth]{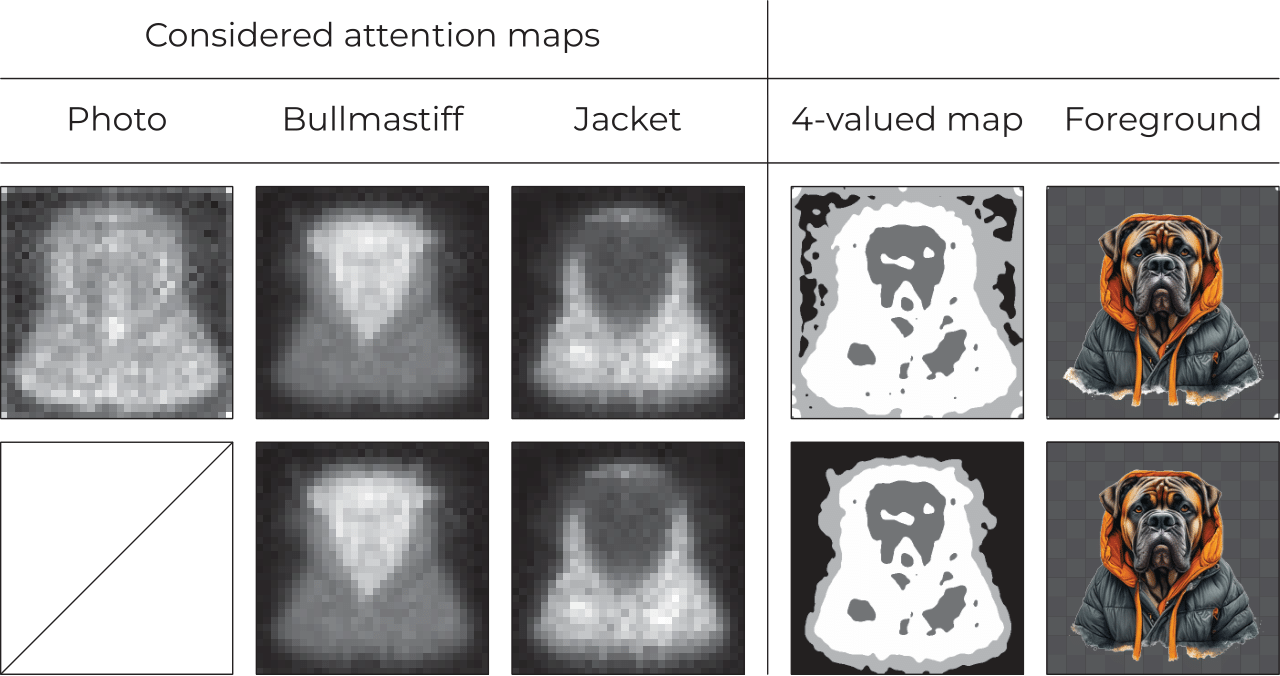}
    \vspace{-0.1cm}
    \caption{Cross-attention map analysis of the prompt \textit{A photo of a bullmastiff with a jacket}. Left to right: maps of the three prompt nouns, candidate region mask computed w/ and w/o the generic noun \textit{photo}, and foreground extraction results.}
    \vspace{-0.7cm}
    \label{fig:comp_sup}
\end{figure}

Once the final RGB image ${\mathbf{x}}_{0}$ is obtained, we isolate the cross-attention maps of the noise estimator relative to the nouns in the input prompt $\mathbf{e}$. The list of nouns $\mathbf{n} = [\mathbf{n}_1, ..., \mathbf{n}_N]$ is obtained by analyzing the prompt with the standard tokenizer from the NLTK library~\cite{bird2009natural} and excluding generic, uninformative nouns such as \textit{image}, \textit{photo}, and \textit{picture}. We release the complete list of excluded nouns in the provided repository.
Empirically, we found that the cross-attention maps associated with these nouns refer to generic areas of the image and not always to the subject, hindering the final quality. This is in line with what is done in~\cite{karazija2023diffusion, tang-etal-2023-daam}, where the authors performed analyses on the cross-attention maps of text-to-image convolutional-attentive diffusion models and avoided spurious correlations between text tokens and feature space by removing this type of nouns. To showcase the effect of the generic nouns exclusion, in~\cref{fig:comp_sup} we visualize the attention maps of different nouns and the effect on the obtained illustration of excluding generic nouns for the prompt \textit{A photo of a bullmastiff with a jacket}. As we can see, the attention maps related to meaningful object nouns (\eg, \textit{dog}, \textit{jacket}) have high activation values in the spatial locations aligned with the specified subjects. In contrast, the attention map of the generic noun~\textit{photo} has high activation values also in border regions. 
In the second-last column, we display the mask of the candidate foreground regions that we use to obtain the guidance masks for a foreground extraction algorithm, namely GrabCut\cite{rother2004grabcut}. In particular, we average the maps of each subject and normalize the values between 0 and 1, obtaining the foreground cross-attention map $\overline{\mathcal{C}\mkern-4mu\mathcal{A}}_{\text{fg}}$ and quantizing it into a mask. For foreground extraction, this mask has four values: background (black), probable background (light gray), probable foreground (white), and sure foreground (dark gray). As we can see, this mask is much cleaner when not considering generic, non-descriptive nouns, and the final illustration is of higher quality. 
In summary, as also shown in~\cite{tang-etal-2023-daam, hertz2022prompt} for U-Net-based diffusion models, cross-attention maps associated with prompt subjects contain high activation values in the foreground regions. We argue that the attention maps of a Transformer-based model exhibit the same behavior. Therefore, we use them to compute the candidate foreground pixels. 

Then, we extract the subject information from the self-attention maps. Inspired by DAAM-I2I~\cite{daam2024}, we use the foreground map $\overline{\mathcal{C}\mkern-4mu\mathcal{A}}_{\text{fg}}$ to merge all the foreground pixels self-attention maps by performing a weighted average, where the weights are the activation values of the foreground maps, normalized in the range $[0,1]$. This map, denoted as $\overline{\mathcal{F}\mkern-4mu\mathcal{F}}_{(\overline{\mathcal{C}\mkern-4mu\mathcal{A}}_{\text{fg}})}$, provides us with the attention scores of each foreground pixel with respect to the other foreground pixels. As explained before, we want to combine the different aspects captured by cross- and self-attention and use both for our coarse $\alpha$ channel estimation, which is given by the normalized sum of the two maps, \ie,
\begin{equation*}\label{eq:alpha_hat}
    \hat{\alpha} = \overline{ \overline{\mathcal{C}\mkern-4mu\mathcal{A}}_{\text{fg}} + \overline{\mathcal{F}\mkern-4mu\mathcal{F}}_{(\overline{\mathcal{C}\mkern-4mu\mathcal{A}}_{\text{fg}})}}. 
\end{equation*}

\tit{Obtaining the RGBA Illustration} 
Once the reverse diffusion process is over, we obtain an image containing the desired subject with no sharp croppings and an estimation of the coarse transparency mask $\hat{\alpha}$. At this point, we need to remove the background to obtain the final illustration.
To this end, we exploit the candidate foreground map $\overline{\mathcal{C}\mkern-4mu\mathcal{A}}_{\text{fg}}$ obtained with the cross-attention maps of the noise estimation Transformer. Specifically, we quantize it into four values to obtain a map whose values indicate whether the corresponding pixel is in the sure background, the probable background, the probable foreground, or the sure foreground. We use the so-obtained mask as input to the perceptual GrabCut~\cite{rother2004grabcut} algorithm, which performs graph optimization by combining color and border information to remove the background from the generated image. We use this algorithm's output foreground mask to clean $\hat{\alpha}$, zeroing the values outside the subject and obtaining our transparency channel $\alpha$. 

As a side note, we remark that one could alter the intensity of the $\alpha$ channel, \ie~augmenting or reducing the opacity of the generated illustration in a semantically meaningful way, by simply introducing a hyperparameter $k$ to obtain 
\begin{equation*}
    \hat{\alpha}' = \text{min}(1, (1+k)\hat{\alpha}),
\end{equation*}
which the user can tune depending on their needs. We set it to 0.5 in all experiments, leaving further exploration for future work.
\section{Experiments and Results}

\tit{Implementation Details} 
In our experiments, we use the pre-trained PixArt-$\Sigma$-512~\cite{chen2024pixartsigma} from HuggingFace~\cite{von-platen-etal-2022-diffusers}. PixArt-$\Sigma$ is based on DiT~\cite{peebles2023scalable} and comprises 28 self- and cross-attention layers, each with 16 attention heads. We use the Euler Discrete scheduler~\cite{karras2022elucidating} with 30 denoising steps. The subject-centering mask has a border of 64 in the pixel space on all image sides. For the attention map quantization into a 4-valued map, we use the percentiles 0.8, 0.3, and 0.1, respectively: for sure foreground, probable foreground, and probable background. 

\tit{Evaluation Setup}
For quantitative evaluation, we collect 3000 of the multi-sentence prompts used to train PixArt-$\Sigma$~\cite{chen2023pixartalpha,chen2024pixartsigma} and consider only the first sentence of the prompt, which is the one describing the general visual content of the image. These have been obtained by running the LLaVA Multimodal Large Language Model~\cite{liu2024visual} on images from the SAM dataset~\cite{kirillov2023segment}\footnote{We release the list of prompts at \url{https://github.com/aimagelab/Alfie.}}. 

We quantitatively evaluate how reliably the pipeline generates illustrations of subjects wholly contained within the image, without crops or cuts. To do so, we consider the image canvas's left, right, top, and bottom borders and, for each one, a margin of 4 pixels. If all the pixels in the margin area have a value greater than 0.8, we consider that border empty. We choose this threshold value because we aim to keep the background uniform and light rather than perfectly white. Then, we compute the percentage of generated images having each of the borders empty (denoted as \textbf{empty}-\{\textbf{l}, \textbf{r}, \textbf{t}, \textbf{b}\}, respectively), as well as the aggregate percentage of images having all four borders empty (denoted as \textbf{empty-a}).  Moreover, we consider the CLIP score~\cite{hessel2021clipscore} (denoted as \textbf{CLIP-S}) to quantify how much the generated RGB images still respect the original prompt. 
Regarding the RGBA images, we remark that we do not evaluate in terms of the FID and KID scores, which are commonly adopted in image generation evaluation. This is because these scores need a large reference set (ground truth RGBA images) possibly compatible with the prompts fed to our model and with the characteristics that we enforce in our images. To the best of our knowledge, such a set with the required characteristics is not available.

Moreover, in line with RGBA image generation works~\cite{zhang2024transparent}, we perform a user study to compare the perceived quality of our illustrations compared to the best-performing alternative method. For a fair comparison, we employ the state-of-the-art ViT-B-based ViTMatte~\cite{yao2024vitmatte} and perform matting over our generation results. This matting method requires an input trimap for sure foreground, unsure regions, and sure background. We provide it by quantizing the same map $\overline{\mathcal{C}\mkern-4mu\mathcal{A}}_{\text{fg}}$ that we obtain in our pipeline with the percentiles 0.8 and 0.3 for sure foreground and unsure regions, respectively. We consider the values under 0.3 as sure background. We also provide qualitative results showcasing the effects of centering, $\alpha$ estimation, and comparison with matting. 

\tit{Considered Variants and Baselines}
As baselines, we consider the base PixArt-$\Sigma$~\cite{chen2024pixartsigma} with the same prompts as we use for our pipeline and a version to which we always append the phrase \textit{on a white background} to the input prompt (we refer to this as \textbf{PixArt-$\Sigma$+suffix}). As for the variants of our approach, we isolate the role of each component by considering the subject centering (which we refer to as \textbf{PixArt-$\Sigma$+centering}) and different methods for obtaining the $\alpha$ channel, namely: the normalized cross-attention heatmap $\overline{\mathcal{C}\mkern-4mu\mathcal{A}}_{\text{fg}}$, the normalized self-attention heatmap $\overline{\mathcal{F}\mkern-4mu\mathcal{F}}_{(\overline{\mathcal{C}\mkern-4mu\mathcal{A}}_{\text{fg}})}$, the normalized combination of cross- and self-attention $\hat{\alpha}$, and the GrabCut-cleaned $\alpha$ (which we refer to as \textit{Alfie}).

\tit{Runtime}
We run all experiments on a 24GB Nvidia RTX 4090 GPU with half precision. Generating an image with the base model (PixArt-$\Sigma$~\cite{chen2024pixartsigma}) takes ${\sim}$15GB of VRAM and ${\sim}$3.15 seconds, increased to ${\sim}$4.13 seconds and ${\sim}$16.7GB of VRAM when we perform the centering. Cleaning the estimate transparency channel $\hat{\alpha}$ with GrabCut~\cite{rother2004grabcut} requires additional ${\sim}$0.36 seconds. 

\vspace{-0.2cm}
\subsection{Results}

\tit{Centering}
We consider the background generation statistics and report it in~\Cref{tab:centering}. We can observe that the PixArt-$\Sigma$+suffix variant is not entirely reliable, as it leads to $\sim$53$\%$ probability of generating the image inside the canvas. Conversely, our approach provides a much higher probability of generating the whole subject, consistently reaching more than 95$\%$. In~\Cref{tab:centering}, we also report the CLIP-S between the generated images and the textual prompt. As we can see, the base model obtains the highest value, our reference score, as the generated images are fully based on the textual prompt, and the generation process is unconstrained.
On the other hand, our target images have implicit characteristics that are not part of the textual caption but rather represent meta-descriptions, \eg, whole-object generation. As expected, the second-best CLIP-S is obtained by the base model with the suffix. In fact, this baseline sometimes (almost half of the times) ignores the suffix, and thus, it is less penalized. While respecting the desired characteristics defined in~\Cref{sec:method}, our generated images achieve CLIP-S values very close to the reference ones. Thus, we can conclude that our inference-time adaptation strategy guides the generation toward the desired characteristics while maintaining adherence to the textual prompt. 
\begin{figure}[t]
    \centering
    \includegraphics[width=\linewidth]{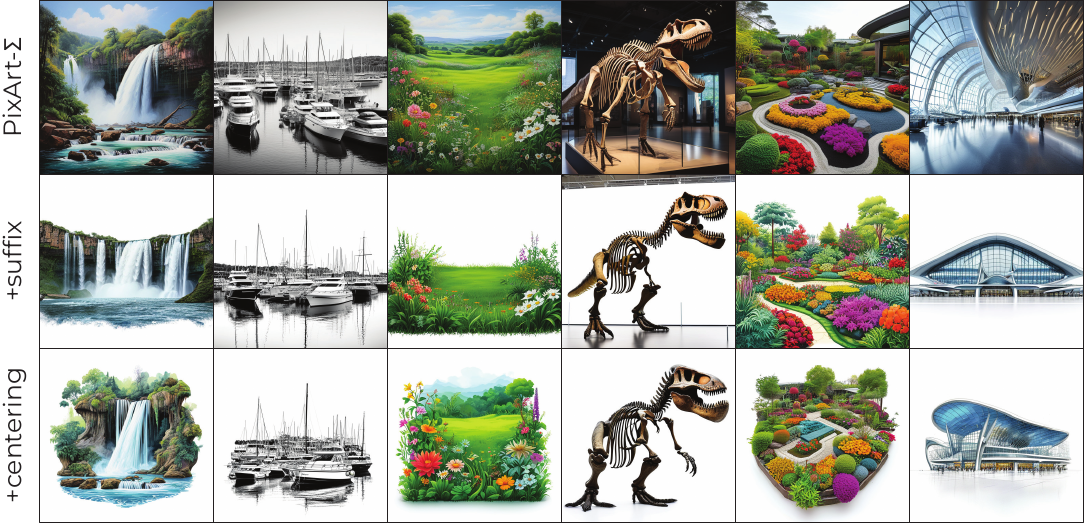}
    \vspace{-0.5cm}
    \caption{Qualitative comparison on the effect of our constrained whole-subject generation compared to meta-descriptions in the input prompt.}\vspace{-0.5cm}
    \label{fig:qualitatives_centering}
\end{figure}
In~\cref{fig:qualitatives_centering}, we showcase the impact of this constraint on the generated images. While the simple appending of the suffix cannot always guarantee whole-subject generation, our method consistently steers the generation towards a correct result and enforces this meta-description on the image. 
\begin{table}[]
    \renewcommand{\arraystretch}{.95}
    \centering
    \setlength{\tabcolsep}{.18em}
     \caption{Quantitative comparison of our approach variants in terms of average probability with which the subjects are generated in the center of the image with no sharp croppings and adherence to the prompt.}\vspace{-0.2cm}
    \begin{tabular}{l c c c c c c}
    \toprule
    & \textbf{empty-a} & \textbf{empty-l} & \textbf{empty-r} & \textbf{empty-t} & \textbf{empty-b} & \textbf{CLIP-S} \\
    \midrule
    \textbf{PixArt-$\Sigma$} & 3.33& 4.33& 4.13& 8.53& 6.06& 31.29 \\
    \textbf{PixArt-$\Sigma$ + suffix}        & 53.10& 62.63& 62.10& 92.53& 84.50& 30.79\\
    \textbf{PixArt-$\Sigma$ + centering}     & 96.50 & 99.27& 99.03& 99.53& 98.00& 30.08\\
    \bottomrule
    \end{tabular}
    \vspace{-3.8em}
    \label{tab:centering}
\end{table}

\begin{figure}[t]
    \centering
    \includegraphics[width=\linewidth]{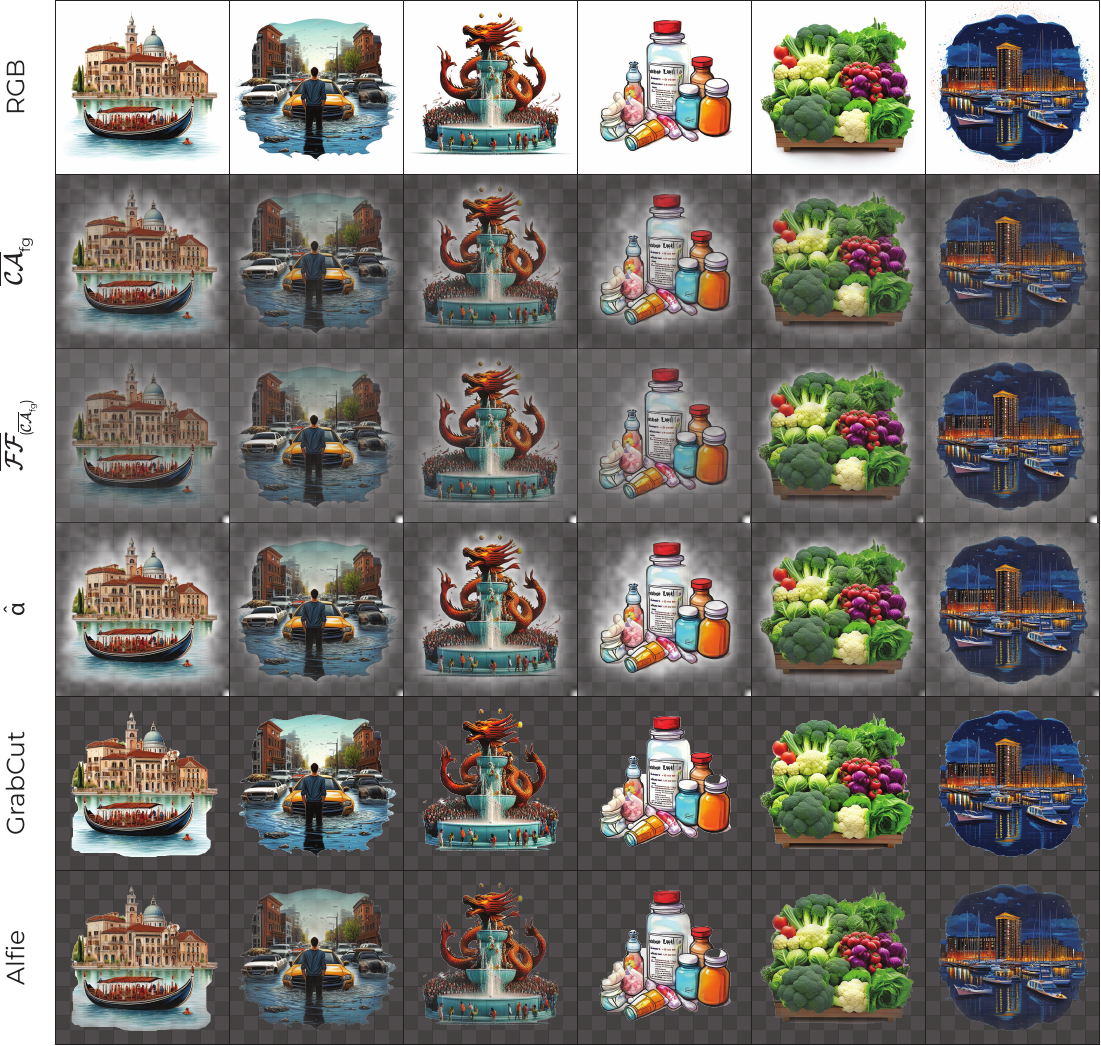}
    \vspace{-0.5cm}
    \caption{Qualitative comparison on different $\alpha$ estimates. Combining self- and cross-attention maps provides the best balance of spatial localization and transparency values, and their cleanup using GrabCut~\cite{rother2004grabcut} (\textit{Alfie}) further increases border precision.}\vspace{-.5cm}
    \label{fig:qualitatives_alpha_estimation}
\end{figure}

\tit{$\alpha$ Channel Estimation}
In~\cref{fig:qualitatives_alpha_estimation}, we show alternative $\alpha$ estimations obtained using different combinations and processing of the attention maps. As we can see, the normalized cross-attention map $\overline{\mathcal{C}\mkern-4mu\mathcal{A}}_{\text{fg}}$ provides coarse localization of the foreground subjects with uniform transparency values. The normalized self-attention map $\overline{\mathcal{F}\mkern-4mu\mathcal{F}}_{(\overline{\mathcal{C}\mkern-4mu\mathcal{A}}_{\text{fg}})}$ is more spatially precise and has a broader range of values, reflecting how much the objects appearance depends on a localized or sparse image region. By combining the two, the resulting map $\hat{\alpha}$ gives a balance between absolute and relative transparency values. Finally, using Grabcut~\cite{rother2004grabcut} for cleanup ensures precise border localization, leading to the best overall results. 

\begin{figure}[t]
    \centering
    \includegraphics[width=\linewidth]{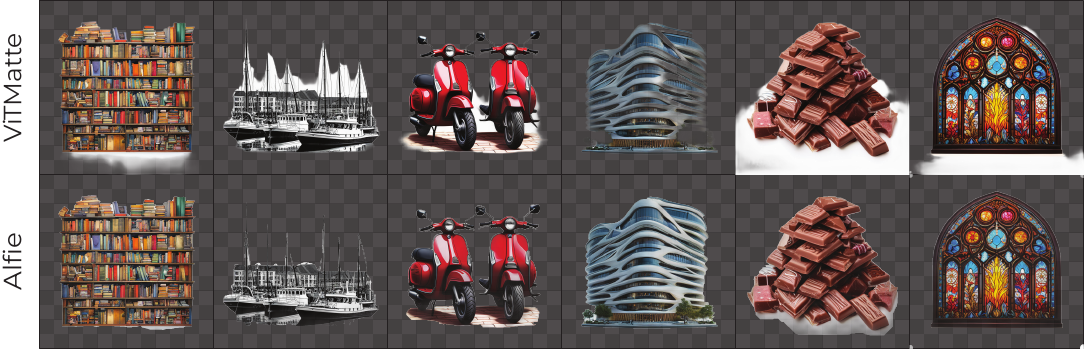}
    \vspace{-0.5cm}
    \caption{Qualitative comparison of our method against ViTMatte~\cite{yao2024vitmatte}. Without training an additional network, we produce reasonable transparency channels.}\vspace{-0.5cm}
    \label{fig:comparison_matting}
\end{figure}

\tit{Comparison with Matting}
We compare our method against matting by considering the generated RGB images with the foreground cross-attention map $\overline{\mathcal{C}\mkern-4mu\mathcal{A}}_{\text{fg}}$ and using it to compute the input trimap for ViTMatte~\cite{yao2024vitmatte}. We run a user study with ten individuals and 100 images randomly sampled from our dataset and ask them "Which of the following RGBA illustrations do you prefer?". We find that users indicated the images from our method $\sim$63\% of the time, demonstrating that we are slightly better than state-of-the-art matting methods while requiring no costs for training. In~\Cref{fig:comparison_matting}, we provide samples obtained by using the two methods. As we can see, our method provides reasonable $\alpha$ channels, comparable to or better than matting, mainly because using the attention maps reduces prediction errors,  discarded image parts, and preserved background areas.

\tit{Compositional Artwork Generation}
We use the images generated by our method as input for Collage Diffusion~\cite{sarukkai2024collage}. This method allows compositional scene generation by taking sequences of layers as input, which define the spatial arrangement and attributes of objects in the scene. The authors combine SDEdit~\cite{meng2021sdedit}, cross-attention manipulation~\cite{balaji2022ediffi}, textual inversion~\cite{gal2022image}, and a proposed extension of ControlNet~\cite{zhang2023adding} to find a balance between harmonization (of the input subjects and the scene) and fidelity (of the produced collage \wrt the different original subjects). The pipeline runs on Stable Diffusion v2-1-base~\cite{rombach2022high}. In~\cref{{fig:composition}}, we show the results on the pipeline with the same input background and three different subjects: an RGBA illustration obtained from Adobe Stock, an image obtained with the base PixArt-$\Sigma$ model, and an image obtained with Alfie. The outputs of Collage Diffusion showcase that the quality obtained by using our method is similar to that achieved with a commercial RGBA illustration. Conversely, when the input is the base PixArt-$\Sigma$ squared illustration, the pipeline tries to incorporate it into the scene by rendering it as a big painting, a plausible but unacceptable result.

\begin{figure}[t]
    \centering
    \includegraphics[width=0.9\linewidth]{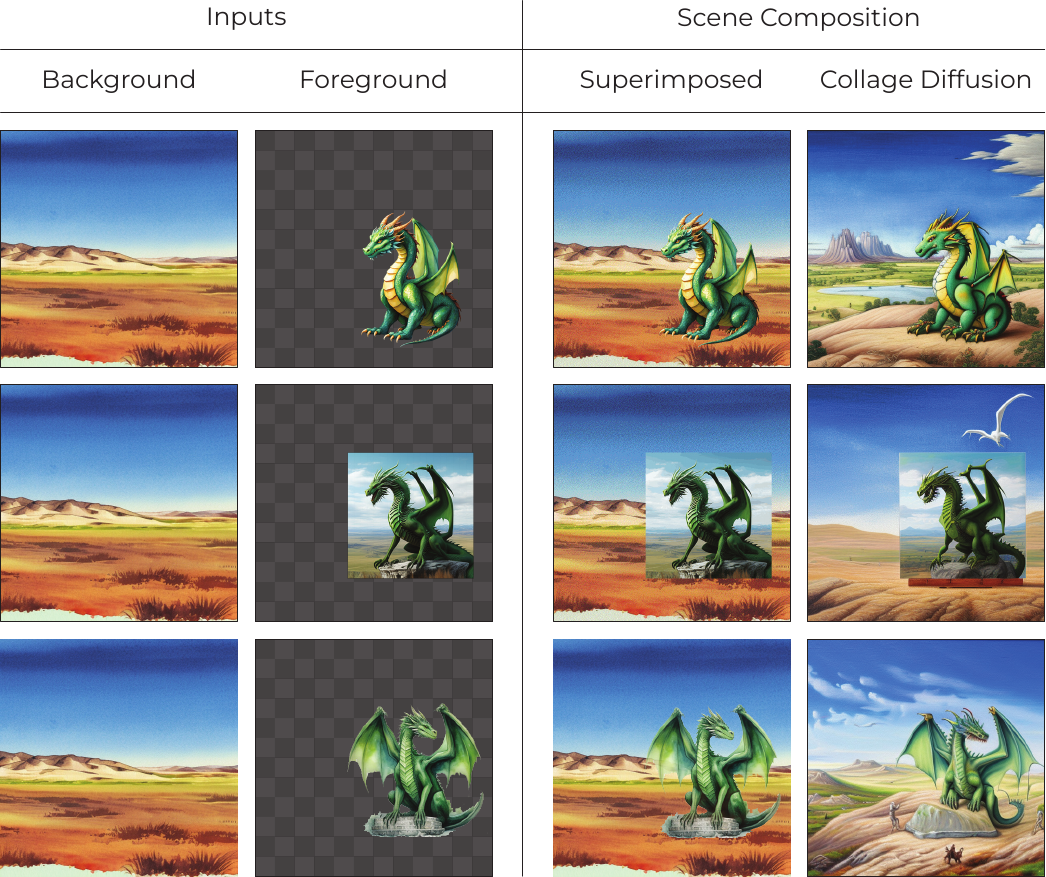}
    \caption{Results of Collage Diffusion~\cite{sarukkai2024collage} with different input foreground images. Top to bottom: a commercial RGBA image, an image generated by the baseline PixArt-$\Sigma$~\cite{chen2024pixartsigma}, and an RGBA illustration obtained with our approach. With our generated illustration, the output is comparable to that obtained with the commercial image, differently from the pipeline run with the image from the baseline model. To generate the background and the foreground, we use the prompts \textit{A steppe landscape} and \textit{A green dragon}, combined for the composite scene into \textit{A green dragon in a steppe landscape}.}\vspace{-0.5cm}
    \label{fig:composition}
\end{figure}

\section{Conclusions}
In this work, we have tackled the task of generating illustrations for visual content, artworks, and visually-rich documents. We have devised a fully automated, prompt-guided pipeline for obtaining high-quality RGBA images that can be used by users or integrated into scene compositions with good results at zero additional training cost.
As the main component of our pipeline, we have exploited the recently proposed Diffusion Transformer paradigm and explored inference-time adaptation strategies for these models. We have quantitatively and qualitatively validated the effectiveness of the proposed approach, which can serve as a starting point for the community to continue working on the tackled task. We hope that the promising results obtained can encourage the research towards the development of similar low-cost strategies for such models.

\section*{Acknowledgement}
This work was supported by the ``AI for Digital Humanities'' project funded by ``Fondazione di Modena'' and the PNRR project Italian Strengthening of ESFRI RI Resilience (ITSERR) funded by the European Union – NextGenerationEU.

\clearpage
\bibliographystyle{splncs04}
\bibliography{main}
\end{document}